\documentclass[10pt,twocolumn,letterpaper]{article}

\pdfoutput=1

\usepackage{graphicx}
\usepackage{caption}
\usepackage{subcaption}
\usepackage{cvpr}
\usepackage{times}
\usepackage{amsmath}
\usepackage{amssymb}
\usepackage{verbatim}
\usepackage{color}

\usepackage[pagebackref=true,breaklinks=true,letterpaper=true,colorlinks,bookmarks=false]{hyperref}

\newcommand{\mypar}[1]{\vspace{-4mm}\paragraph{#1}}

 \cvprfinalcopy 

\def\cvprPaperID{1816} 

\ifcvprfinal\fi
\begin{document}

\title{Situational Object Boundary Detection}

\author{J.R.R. Uijlings\\
University of Edinburgh\\
\and V. Ferrari \\
University of Edinburgh\\
}

\maketitle
\thispagestyle{empty}

\begin{abstract}

Intuitively, the appearance of true object boundaries varies from image to image. Hence the usual
monolithic approach of training a single boundary predictor and applying it to all images regardless
of their content  is bound to be suboptimal.  In this paper we
therefore propose \emph{situational object boundary detection}: We first define a variety of
situations and train a specialized object boundary detector for each of them using~\cite{Dollar13}.
Then given a test image, we classify it into these situations using its context, which we model by
global image appearance. We apply the corresponding situational object boundary detectors, and fuse
them based on the classification probabilities.  In experiments on 
ImageNet~\cite{Russakovsky14}, Microsoft COCO~\cite{lin14eccv}, and Pascal VOC 2012
segmentation~\cite{everingham14ijcv} we show that our situational object boundary detection gives
significant improvements over a monolithic approach. Additionally, our method substantially
outperforms~\cite{hariharan11iccv} on semantic contour detection on their SBD dataset.


\end{abstract}

\vspace{-.3cm}
\section{Introduction}

\begin{figure}[thpb]
    \vspace{-.2cm}
    \hspace{-.6cm}
    \centering
    \includegraphics[width=1.1\linewidth]{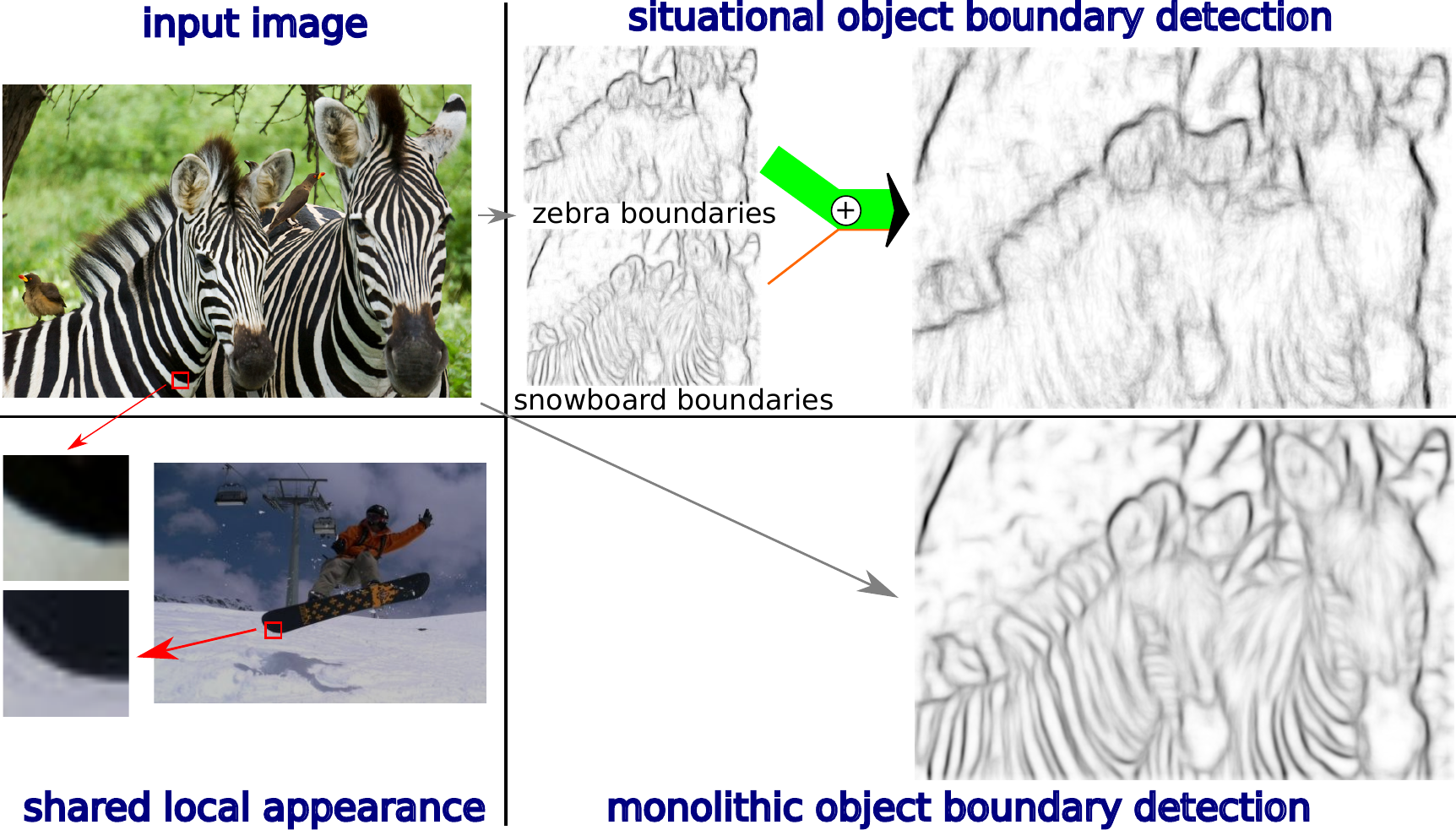}
    \caption{\emph{Monolithic vs situational object boundary detection. Black-white transitions indicate
        an object boundary for the snowboard, but are false object boundaries for a zebra. This
        ambiguity cannot be resolved by a monolithic detector. In contrast, by training class
    specific object boundary detectors and classifying the image as a zebra, we correctly ignore
most of the stripes.}}
    \label{figZebra}
    \vspace{-4mm}
\end{figure}

Most methods for object boundary detection are monolithic and use a single predictor to
predict all object boundaries in an image~\cite{Arbelaez11,Dollar13,Lim13} regardless of the image
content. But intuitively, the appearance of object boundaries is dependent on what is depicted in
the image. For example, black-white transitions are often good indicators of object boundaries,
unless the image depicts a zebra as in Figure~\ref{figZebra}. Outdoors, the sun may cast shadows which
create strong contrasts that are not object boundaries, while similar colour contrasts in an indoor
environment with diffuse lighting may be caused by object boundaries. Furthermore, not all objects
are equally important in all circumstances: one may want to detect the boundary between a snowy
mountain and the sky in images of winter holidays, while ignoring sky-cloud transitions in images
depicting air balloons, even though such boundaries may be visually very similar. These examples
show that one cannot expect a monolithic predictor to accurately predict object boundaries in all
situations. 

In this work we recognize the need for different object boundary detectors in different situations:
first we define a set of situations and pre-train object boundary detectors for each of them. For a
test image, we classify which situations the image depicts based on its context, modelled by
global image appearance. Then we apply the appropriate set of object boundary detectors. Hence
conditioned on the situation of an image we choose which object boundary detectors to run. We call
this Situational Object Boundary Detection.  

One important question is how to define such situations. Since the appearance of object boundaries
are for a large part dependent on the object class, one natural choice is to use each object class
as a single situation. This results in \emph{class specific} object boundary detectors, which can
deal for example with the zebra in Figure~\ref{figZebra}. However, object boundaries are also
determined by the object pose and the background or context of the image. Since this can vary within
a single object class, we propose to cluster images of a single class into subclasses based on
global image appearance. This leads to \emph{subclass specific} object boundary detectors. Finally,
one can imagine that the context of the image itself determines what kind of object boundaries to
expect. For example, one can expect cow-grass boundaries in the countryside and street-car
boundaries in the city. Therefore we cluster images based on their global image appearance,
which results in \emph{class agnostic} object boundary detectors.  Hence we experiment with three
types of situations: \emph{class specific}, \emph{subclass specific}, and \emph{class agnostic}. 

Obviously, situational object boundary detection requires more training data than a monolithic
approach. Therefore we cannot use the standard BSD500~\cite{Arbelaez11} dataset of 500 images for
our evaluation. Instead, we evaluate on three larger datasets: Pascal VOC 2012
segmentation~\cite{everingham14ijcv}, Microsoft COCO~\cite{lin14eccv}, and part of
ImageNet~\cite{Russakovsky14}. Microsoft COCO is two orders of magnitude larger than BSD500. For
ImageNet we train from segments which are created in a semi-supervised fashion by Guillaumin et
al.~\cite{guillaumin14ijcv}.

Additionally, our class-specific situational object boundary detectors can also be applied to 
semantic contour detection, the task of predicting class-specific object
boundaries~\cite{hariharan11iccv}. We compare with~\cite{hariharan11iccv} on their SBD 
dataset.

\section{Related Work}

\paragraph{Manually defined predictors.}

Early work on object boundary detection aimed to manually define local filters to generate edges
from an image. In these works, convolutional derivative filters are applied to find local image
gradients~\cite{Duda73,Prewitt70,Roberts65} and their local maximum~\cite{Canny86,Marr80}. 


\mypar{Trained predictors.}

But object boundaries arise from a complex combination of local cues.  Therefore more recent
techniques resort to machine learning and datasets with annotated object boundaries: Martin et
al.~\cite{Martin04} compute local brightness, colour, and texture cues, which they combine using a
logistic model. Both Mairal et al.~\cite{Mairal08} and Prasad et al.~\cite{Prasad06} use
RGB-features from local patches centred on edges found by the canny edge detector~\cite{Canny86},
which they classify as true or false positives. Doll\'ar et al.~\cite{Dollar06} use boosted decision
trees to predict if the centre label of an image patch is an object boundary or not. Lim et
al.~\cite{Lim13} use Random Forests~\cite{Breiman01} to predict sketch tokens, which are object
boundary \emph{patches} generated by k-means clustering. Doll\'ar and Zitnick~\cite{Dollar13}
proposed structured random forests, which use object boundary patches as structured output labels
inside a random forest. Their method is extremely fast and yields state-of-the-art
results. We build on~\cite{Dollar13} in our paper.



\mypar{Domain specific predictors.}

Some works that use machine learning to predict object boundaries observed that this enables tuning
detectors to specific domains. Doll\'ar et al.~\cite{Dollar06} showed qualitative examples of
domain-specific detectors for finding mouse boundaries in a laboratory setting and detecting streets
in aerial images. Both~\cite{Mairal08} and~\cite{Prasad06} used class-specific object boundary
detectors for boundary-based object classification. Whereas in all these cases the domain was
predefined, in this work we automatically choose which object boundary detector to apply at runtime.

\mypar{Semantic contour detection.}

Like~\cite{Mairal08} and~\cite{Prasad06}, Hariharan et al.~\cite{hariharan11iccv} addressed
class-specific object boundary detection. They call this `semantic contour detection' and create the
SBD benchmark to directly evaluate this task. Their method combines a monolithic object boundary
detector (gPb~\cite{Arbelaez11}) with object class detectors (Poselets~\cite{bourdev10eccv}).  Since
the class-specific version of our situational object boundary detection can readily be applied to
semantic contour detection, we compare to~\cite{hariharan11iccv} in
Section~\ref{cptSemanticContourDetection}.

\mypar{Globally constrained predictors.}

Instead of predicting boundaries only at a local level, Arbel\'aez et al.~\cite{Arbelaez11} cast the
problem into a global optimization framework capturing non-local properties in the spirit of
Normalized Cuts~\cite{Shi00}. In this paper we use the global image appearance to determine the set
of local object boundary predictors to use. In this sense, the global appearance of the image
restricts our algorithm to a limited set of expected object boundaries.

\mypar{Contextual guidance.}

Context, as modelled by global image appearance, has been successfully used to guide a variety of
computer vision tasks. Torralba et al.~\cite{torralba03ijcv} showed that global image features
effectively constrain both the object class and its location, which is frequently
used in object localisation (e.g.~\cite{everingham14ijcv,Felzenszwalb10,harzallah09iccv}). Boix
et al.~\cite{boix12ijcv} do semantic segmentation by region prediction, where the global
image appearance enforces a consistency potential in their hierarchical CRF.  Liu et
al.~\cite{liu09cvpr} perform semantic segmentation through label transfer. Given a test image, they
retrieve nearest neighbours from a pixel-wise annotated dataset using global image appearance. After
region alignment, they transfer labels to the test image. In this paper
we use context modelled by global image features to select those object boundary detectors that
correspond to the situation depicted in the image.

\section{Method}

\subsection{Situational Object Boundary Detection}\label{cptMethodSituational}

\begin{figure}[t]
    \vspace{-.7cm}
    \hspace{-0.7cm}
    \includegraphics[width=1.1\linewidth]{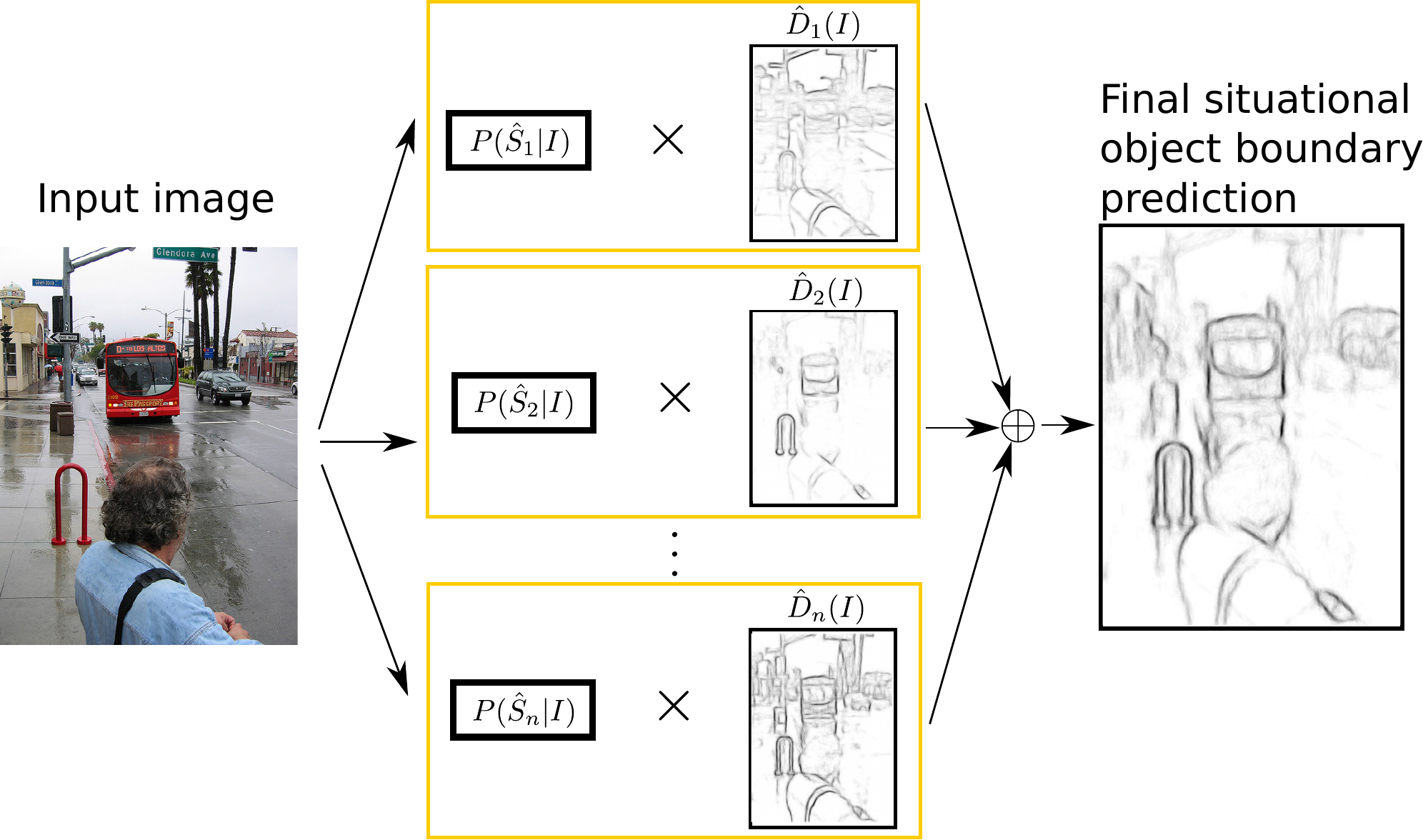}
    \caption{\emph{Overview of situational object boundary detection. For each
        situation there is a specialised boundary detector $\hat{D}_j$ which we apply
        by $\hat{D}_j(I)$. The specialised predictions vary greatly and are combined into a
        final prediction using Equation~\eqref{eqMainCropped}.}}
        \label{figMethod}
        \vspace{-4mm}
\end{figure}

Our main idea is visualized in Figure~\ref{figMethod}. For each specific situation, one can train a
specialized object boundary detector. Given a test image, one then only needs to apply those
boundary detectors which best fit its situation. Intuitively, the global image appearance can help
distinguish the local appearance of true object boundaries from edges caused by other phenomena.  

Formally, let $\mathcal{D} = \{D_1, \ldots, D_k\}$ be a set of $k$ trained object boundary detectors
for a corresponding set of $k$ situations $\mathcal{S} = \{S_1, \ldots, S_k\}$.  Applying the $j$-th
detector $D_j$ to image $I$ gives the boundary prediction $D_j(I)$.
We write the probability that image $I$ corresponds to situation $S_j$ as $P(S_j|I)$, which we
obtain using global image classification as explained in
section~\ref{cptImageClassification}. Now we get the final object boundary prediction
$\mathbf{D}(I)$ by:
\begin{equation}
    \mathbf{D}(I) = \sum_{j=1}^k P(S_j|I) \cdot D_j(I)
    \label{eqMain}
\end{equation}
Of course, we do not need to apply all object boundary detectors to the image since $P(S_j|I)$ is
likely to be small for most situations $j$. To reduce computational costs we take the top few $n \ll k$
situations for which $P(S_j|I)$ is highest. Formally, let $\hat{\mathcal{S}} = \{\hat{S}_1, \ldots,
\hat{S}_k\}$ be an ordered set for a specific image $I$ such that $P(\hat{S}_i|I) > P(\hat{S}_j|I)$
for all $i < j$. Let $\mathcal{\hat{D}}$ be the set of boundary detectors corresponding to
$\hat{S}$. Then the final object boundary prediction is obtained by:
\begin{equation}
    \vspace{-1mm}
    \mathbf{D}(I) = \frac{1}{Z} \sum_{j=1}^n P(\hat{S}_j|I) \cdot \hat{D}_j(I)
    \label{eqMainCropped}
\end{equation}
where $Z=\sum_{j=1}^{n} P(\hat{S}_j|I)$ is a normalizing factor ensuring that the values of the
predicted boundaries are comparable for different $n$ and across images.

We have two choices for $n$: either we fix $n$ or we take $n$ such that $Z > m$ for
a specific probability mass $m$. We determine the best solution experimentally in
section~\ref{cptResultsImageNet}. 

\subsection{Situations}\label{cptSituations}

\begin{figure*}[tbhp]
    \vspace{-.7cm}
    \centering
    \begin{subfigure}[h]{0.32\textwidth}
        \includegraphics[width=\textwidth]{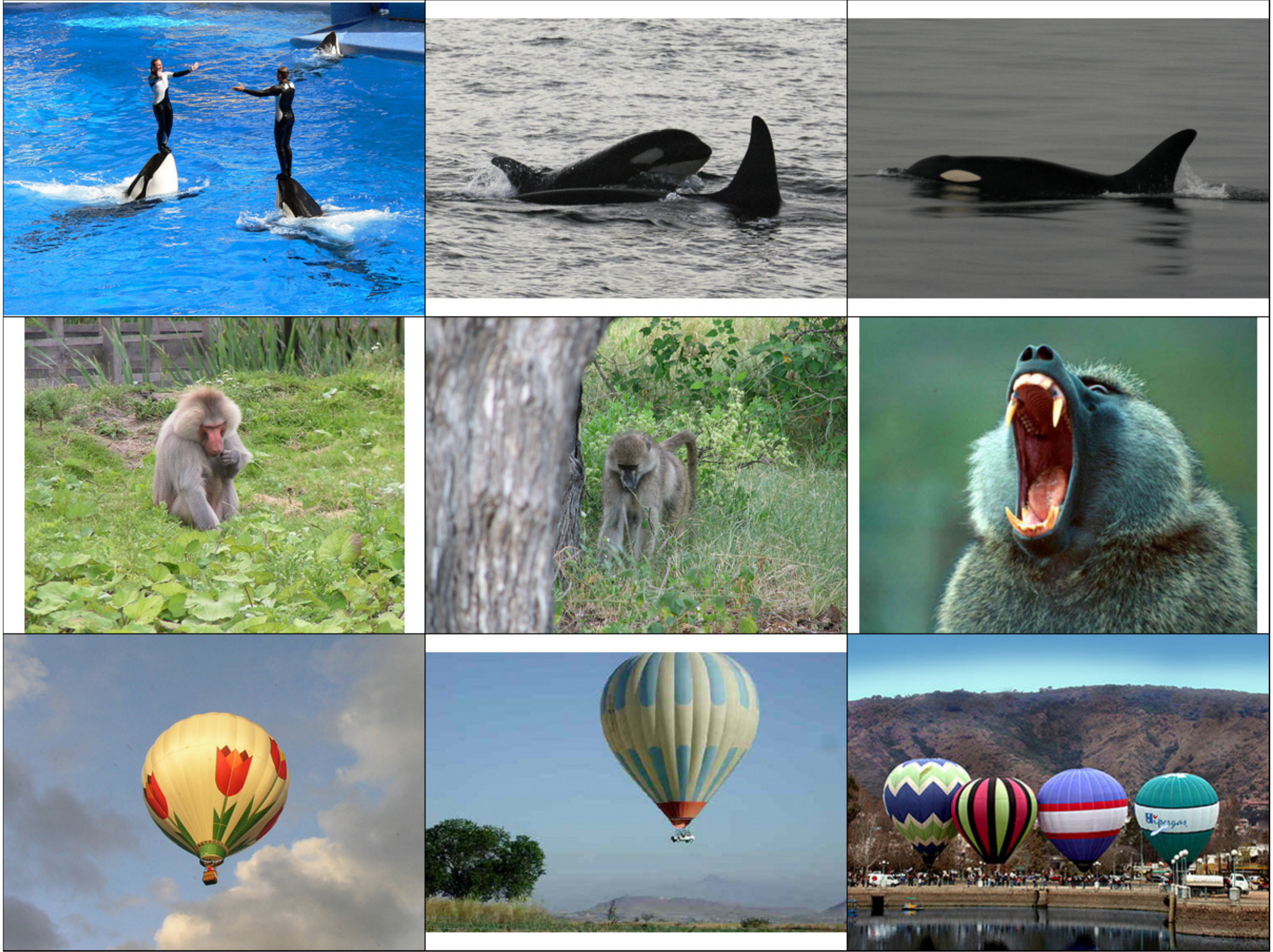}
    \caption{Class specific}
    \label{figClassSpecific}
\end{subfigure}
\hspace{.1cm}
    \begin{subfigure}[h]{0.32\textwidth}
        \includegraphics[width=\textwidth]{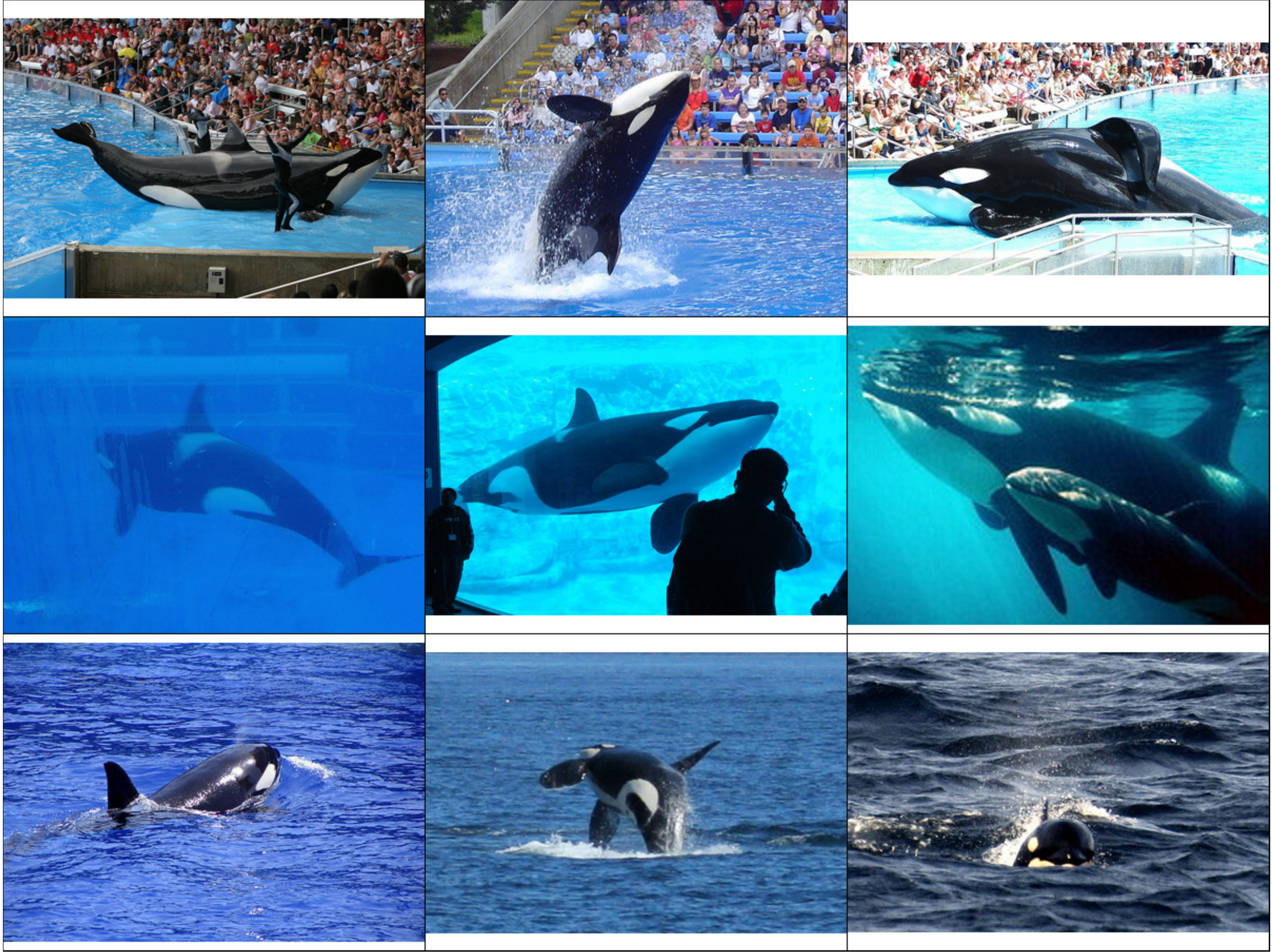}
    \caption{Subclass specific}
    \label{figSubclassSpecific}
\end{subfigure}
\hspace{.1cm}
    \begin{subfigure}[h]{0.32\textwidth}
        \includegraphics[width=\textwidth]{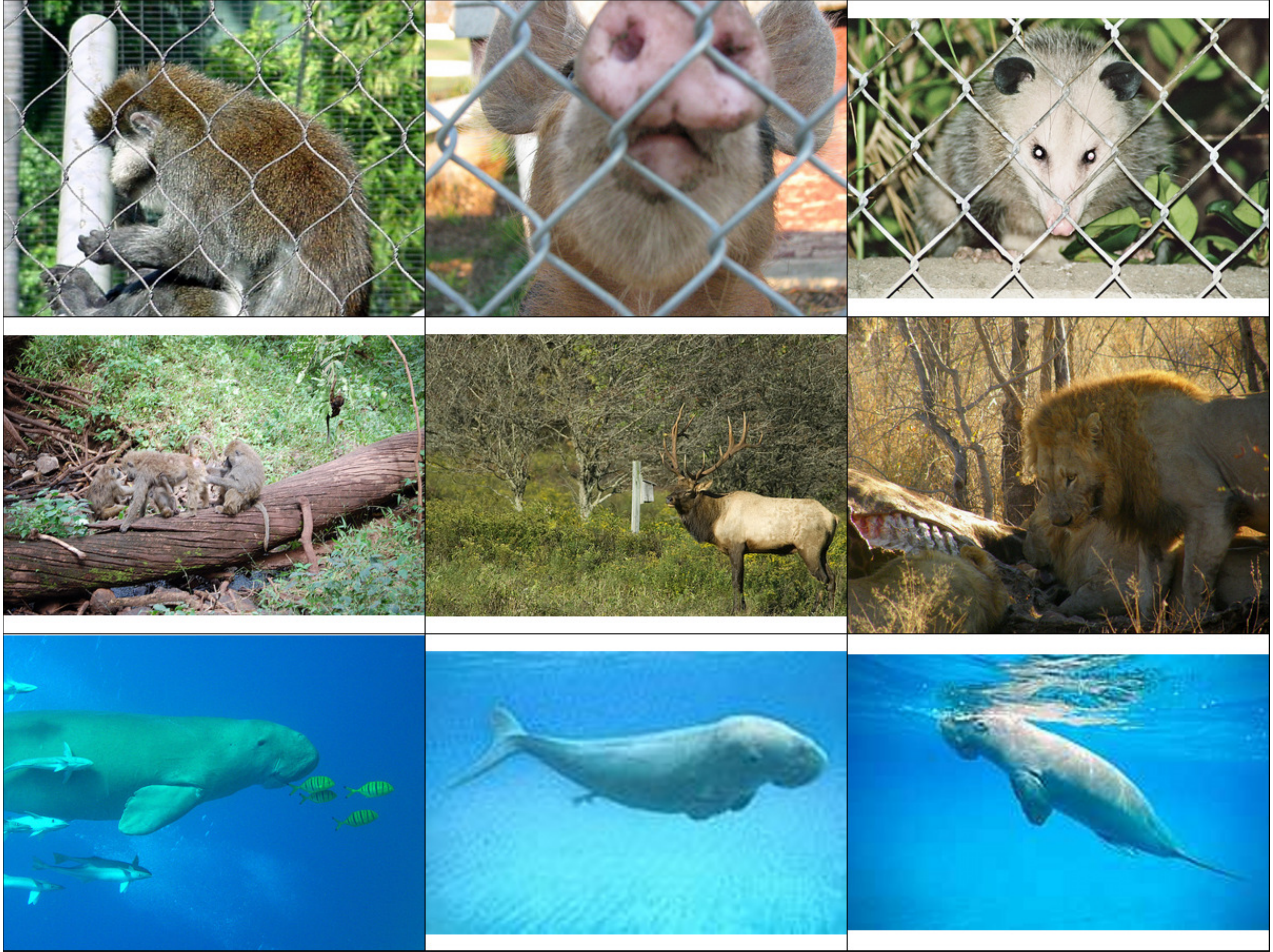}
    \caption{Class agnostic}
    \label{figAppearanceBased}
\end{subfigure}
\caption{\emph{Visualisation for the three types of situations used in this paper. Each row per
    subfigure depicts three example images of a single situation on ImageNet
    (Section~\ref{cptResultsImageNet}).  Figure~\ref{figClassSpecific} shows class
    specific situations, where each situation is a single object class.
    Figure~\ref{figSubclassSpecific} show subclass specific situations, beneficial for
    classes with significant context or
    pose variation such as the killer-whale. Finally, Figure~\ref{figAppearanceBased} shows class
    agnostic situations, which results in contextually similar clusters, some 
containing predominantly images of a single class.}}
\label{figSituations}
\vspace{-.5cm}
\end{figure*}

For situational object boundary detection to work, the key is to define proper situations.  We
propose three ways to define our situations as visualised in Figure~\ref{figSituations}: \emph{class
specific}, \emph{subclass specific}, and \emph{class agnostic}.

\mypar{Class specific.} 

As the term already says itself, object boundaries are caused by the presence of an object. A
logical way to define a situation is therefore to use class specific situations, leading to class
specific boundary detection. We use class labels from the dataset to obtain these
situations.

Class specific situations constrain the appearance of object boundaries in two ways. Most
importantly, instances of the same class tend to have similar appearance: in
Figure~\ref{figClassSpecific} the boundaries of a baboon are all a specific type of fur, while air
balloons have a characteristic oval shape. Second, objects often occur in similar contexts:
killer-whales are mostly in the water while balloons are often in the air. If both the context and
object class is the same, there is little variation in the appearance of object boundaries and one
can learn an object boundary detector which is sensitive to these specific object boundaries.

\mypar{Subclass specific.} 

For some classes, its instances are depicted in a variety of contexts, poses, and from a variety of
viewpoints, which can significantly influence the appearance of the object boundaries. Take for
example the killer-whale in Figure~\ref{figSubclassSpecific}. Photographed in the wild the object
boundaries are only caused by water-whale transitions, while in a whale-show object boundaries can
also be caused by crowd-whale transitions. Furthermore, spurious edges caused by the crowd should
not yield object boundaries here. Additionally, a viewpoint from within the water or from above the
water causes the object boundaries to be very different due to colour changes and absence/presence
of foaming water or waves.  Pose may also affect object boundary appearance: a sleeping, curled-up
cat has much smoother boundaries than a playing cat.

We create subclass specific situations by taking all images of a certain class, model their global
image appearance as described in Section~\ref{cptImageClassification}, and apply k-means
clustering.

\mypar{Class agnostic.} 

Finally, the appearance of object boundaries may be more influenced by context than by the object
class itself. For example, as visualised in Figure~\ref{figAppearanceBased}, photographs taken
through a fence yield spurious edges which are not object boundaries. Detecting such situation
allows for using an object boundary detector which ignores edges from this fence.  Furthermore,
various object classes occur in similar contexts and share characteristics. Indeed, the second row
shows furry animals in a forest environment, giving rise to a similar appearance of object
boundaries.

Therefore the last situation type we consider is class agnostic. We ignore all class labels and
cluster all images of the training set using k-means on global image appearance. As shown
in Figure~\ref{figAppearanceBased}, this leads to clusters of objects in similar contexts, some
with predominantly instances of a single class.

\subsection{Image Classification}\label{cptImageClassification}

For each situation $S_j \in \mathcal{S}$ we need to predict $P(S_j|I)$. We do this using either
Bag of Visual Words~\cite{Csurka04,Sivic03} or Convolutional Neural Net (CNN)
features~\cite{Krizhevsky12}. 

\mypar{Bag of Visual Words.} We extract SIFT descriptors~\cite{Lowe04} of $16 \times 16$ pixels on a
dense regular grid~\cite{Jurie05} at every 4 pixels using~\cite{Uijlings10}. We use PCA to reduce
SIFT to 84 dimensions. We train a GMM with diagonal covariance of 64 clusters. We then create Fisher
Vectors following~\cite{Perronnin10}: we use derivatives only with respect to the means and standard
deviations of the GMM. Vectors are normalized by taking the square root while keeping the sign,
followed by L2 norm. We use a spatial partitioning~\cite{Lazebnik06} using the whole image and a
division into three horizontal regions (e.g.~\cite{Uijlings10}). The final Fisher representation has
43008 dimensions.

\mypar{CNN features.} We use the publicly available software for
deep Neural Networks of Jia et al.~\cite{Jia14}. Instead of training a specialized network for each
dataset, we choose the more flexible option of using a pre-trained network, removing the final
classification layer, and using the last layer as global image appearance features. This was shown
to yield excellent features by e.g.~\cite{donahue13decaf,Girshick14,razavian14cvpr}.

In particular, we use the pre-trained network modelled after Krizhevsky~\cite{Krizhevsky12} that
comes with~\cite{Jia14}, trained on the training set of the ILSVRC classification
task~\cite{Russakovsky14}. This network takes as input RGB images rescaled to $227 \times
227$ pixels. It consists of five convolutional layers, two fully connected layers, and a final
classification layer which we discard. Hence we use the outputs of the 7-th layer as CNN features,
yielding features of 4096 dimensions.

\mypar{Classification.} For both the Fisher Vectors and CNN features, we train linear SVMs with
Stochastic Gradient Descent using~\cite{Vedaldi10VLFeat}. We use cross-validation to optimize the
slack-parameter and, following~\cite{Akata14}, to optimize the relative sampling frequency of
positive examples.

\subsection{Boundary Detector}\label{cptStructuredEdgeForests}

As boundary detector we use the Structured Edge Forests of Doll\'ar and Zitnick~\cite{Dollar13}, as
these are extremely fast and yield state-of-the-art performance. Using their standard settings,
their detector predicts $16 \times 16$ pixel boundary masks from $32 \times 32$ pixel local image
patches. From each local image patch a variety of colour and gradient features is extracted. They
train a random forest directly on the structured output space of segmentation masks: at each node
they sample 256 random pixel pairs and perform binary tests checking if both pixels come from the
same segment. The resulting 256 dimensional vector is reduced to a single dimension using PCA, where
its sign is used as a binary label. This allows for the calculation of information gain as usual.


Unless mentioned otherwise, we use their framework with standard settings except for the number of
training patches. We lower these from 1 million to 300,000 resulting in similar performance
as shown in Section~\ref{cptResultsImageNet}.

\section{Results}\label{cptResults}

In Section~\ref{cptResultsImageNet} to~\ref{cptResultsPascal}, we evaluate our method on object
boundary detection on ImageNet~\cite{Russakovsky14}, Microsoft
COCO~\cite{lin14eccv}, and Pascal VOC 2012 segmentation~\cite{everingham14ijcv}.  We use the
evaluation software of~\cite{Martin04}, average results over all images and report precision/recall
curves, precision at 20\% and 50\% recall, and average precision (AP).

In Section~\ref{cptSemanticContourDetection}, we evaluate our method on semantic contour detection
on the SBD database~\cite{hariharan11iccv} using their evaluation software and report average
precision (AP).

\subsection{ImageNet}

\paragraph{Dataset.}\label{cptResultsImageNet}

While ImageNet has no manually annotated object boundaries, Guillaumin et al.~\cite{guillaumin14ijcv}
obtained good segmentations using a semi-supervised segmentation transfer strategy, applied to increasingly difficult image subsets. As our training set, we use their most reliable segmentations
created from bounding box annotations. As test set, we use the ground-truth segmentations collected by~\cite{guillaumin14ijcv}. 

To keep evaluation time reasonable we randomly sample 100 classes
from the set of~\cite{guillaumin14ijcv}. This results in 23,457
training and 1,000 test images. Since each image is annotated with one object class,
this experiment evaluates only boundaries of that class.

\mypar{Number of situations.}
For subclass specific situations, we choose to cluster classes into
10 subclasses, yielding 1000 situations.
For good comparison, we choose to also have the same
number of 1000 class agnostic situations.

\mypar{Number of detectors at test time.}

We now establish the number of object boundary detectors to apply to get optimal
performance using Equation~\eqref{eqMainCropped}.  Table~\ref{tabSituations} shows results when
varying $n$ for subclass specific object boundary detection (other situations yield similar
results). As can be seen, starting from  $n=5$ results saturate for both methods. Looking at
the probability mass $Z$, at $n=5$ it is 61\% for Fisher vectors and 71\% for CNN features.  However, 
$Z$ greatly differs per image. Hence for stable and efficient computation time with optimal
performance, we fix $n=5$ random forest detectors (of 8 trees) for all subsequent experiments.

\begin{table}[t]
    \vspace{-7mm}
    \begin{footnotesize}
    \centering
    \begin{tabular}{|l|c|c|c|c|}
        \hline
        & $n=1$ & $n=3$ & $n=5$ & $n=25$ \\
        \hline
        $Z$ - CNN \hspace{0.5mm} - subclass specific & 47\% & 65\% & 71\% & 85\% \\
        $Z$ - Fisher - subclass specific & 29\% & 51\% & 61\% & 79\% \\
        \hline
        AP - CNN \hspace{0.5mm} - subclass specific & 0.274 & 0.289 & 0.296 & 0.295 \\
        AP - Fisher - subclass specific & 0.267 & 0.283 & 0.290 & 0.291 \\
        \hline
        AP - Monolithic & 0.258 & 0.259 & 0.260 & 0.260 \\
        \hline
    \end{tabular}
    \caption{\emph{Influence of number of situational object boundaries detectors applied at test
    time. Results saturate in average precision (AP) after applying 5 object boundary detectors.}}
    \label{tabSituations}
    \end{footnotesize}
    \vspace{-1mm}
\end{table}

\begin{table}\begin{footnotesize}
    \centering
\begin{tabular}[h]{|l|c|c|c|}
    \hline
    & precision at & precision at & average \\
    & 20\% recall & 50\% recall & precision \\
    \hline
    monolithic & 0.382 & 0.282 & 0.260 \\
    \hline
    CNN   - class specific & 0.435 & 0.311 & 0.289 \\
    CNN   - subclass specific & 0.451 & 0.317 & 0.296 \\
    CNN   - class agnostic & 0.446 & 0.315 & 0.295 \\
    \hline
    Fisher - class specific & 0.426 & 0.305 & 0.283 \\
    Fisher - subclass specific & 0.442 & 0.312 & 0.290 \\
    Fisher - class agnostic & 0.429 & 0.307 & 0.284 \\
    \hline
    GT - class specific & 0.433 & 0.311 & 0.290 \\
    \hline
    monolithic - CNN enhanced & 0.385 & 0.278 & 0.259 \\
    \hline
\end{tabular}
    \caption{\emph{Results on ImageNet show that situational object boundary detection significantly outperforms
    a monolithic strategy.}}
    \label{tabImageNet}
\end{footnotesize}
        \vspace{-3.5mm}
\end{table}

\mypar{Baseline.}

Our baseline (\emph{monolithic}) is a single monolithic detector. However, for a fair comparison 
our baseline should be trained on the same number of training patches and use the same number of
decision trees. This is equivalent to training multiple monolithic detectors~\cite{Dollar13}. As
shown in Table~\ref{tabSituations}, results are affected little by training more monolithic
detectors, and stabilize at $n=5$ at 0.260 AP. 

We also trained a random forest with the recommended 1M training examples~\cite{Dollar13}
instead of 300k. This yields 0.262 AP. Since this is not
significantly different, for consistency of all experiments we choose as baseline $n=5$ random
forests trained on 300k examples per tree.







\mypar{Situational Object Boundary Detection.}

\begin{figure}[t]
    \vspace{-.8cm}
    \centering
    \includegraphics[width=\linewidth]{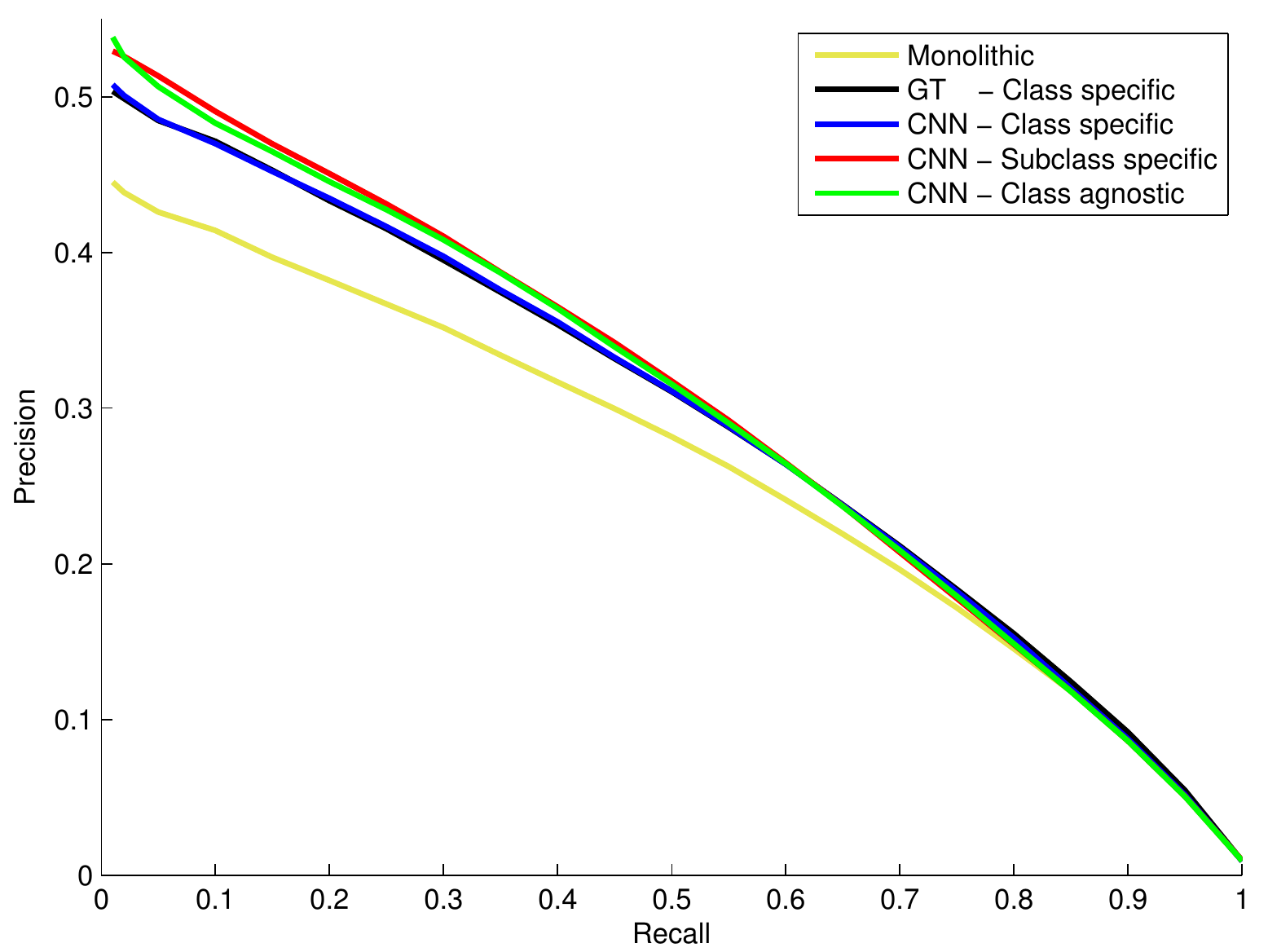}
    \caption{\emph{Performance of object boundary detection on ImageNet. Situational object
        boundary detection significantly outperforms monolithic. The black
    line is occluded by the blue.}}
\label{figImageNet}
\vspace{-.4cm}
\end{figure}

Figure~\ref{figImageNet} and Table~\ref{tabImageNet} show that situational object
boundary detection significantly outperforms the monolithic approach. Using CNN features, at 20\%
recall, the precision for monolithic is 0.38, while it is respectively 0.44, 0.45, and 0.45 for class
specific, subclass specific, and class agnostic situations. 

\begin{figure*}[htpb]
    \vspace{-0.8cm}
    \hspace{-.7cm}\includegraphics[width=1.096\linewidth]{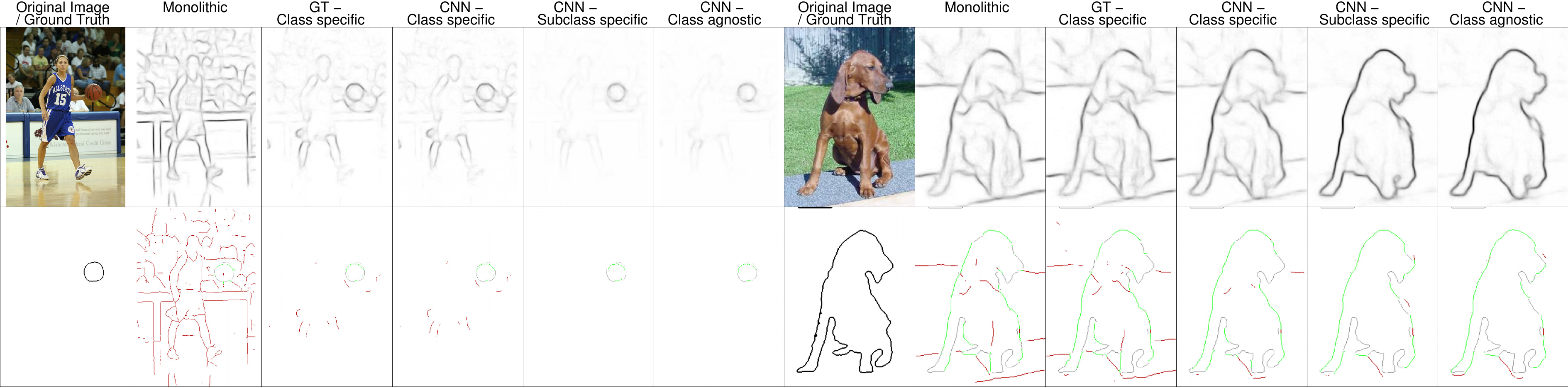}
    \hspace{6cm}\includegraphics[width=\linewidth]{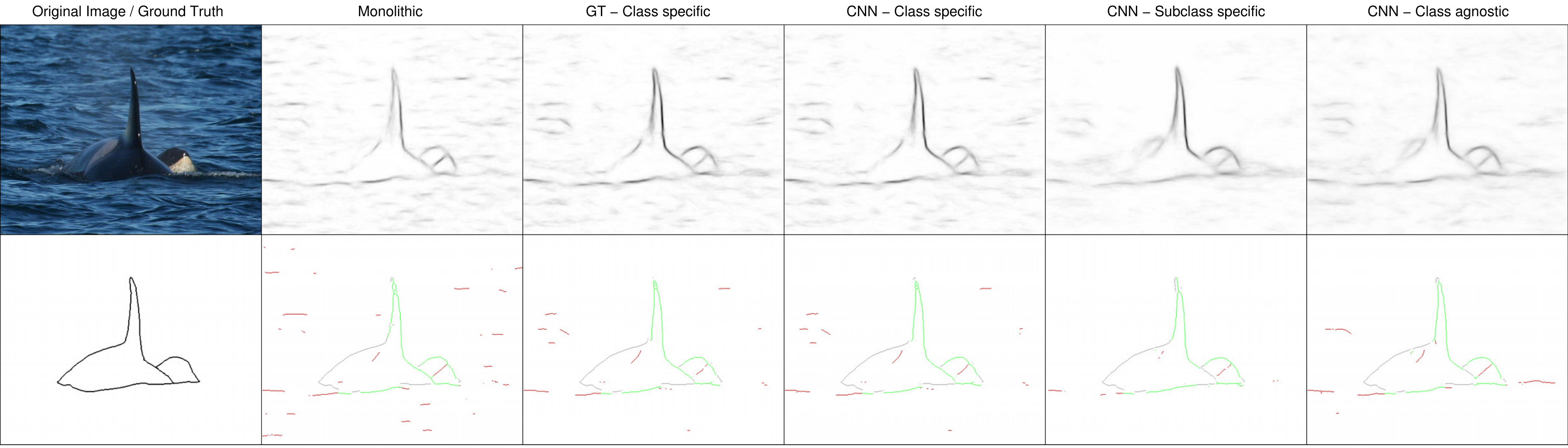}\hspace{-10cm}
    \caption{\emph{Qualitative comparison for monolithic versus situational object boundary detection. The
        upper row shows object boundary predictions. The lower row shows the ground truth boundaries
        and evaluation at 50\% recall, with true positives in green, false positives in red, and
    undetected boundaries in grey. Monolithic boundary detection fires on many false object
boundaries caused by the background and internal boundaries, while situational object boundary detection
focuses much better on the boundaries of the object of interest.}}
    \label{figExampleImageNet}
    \vspace{-5mm}
\end{figure*}

Figure~\ref{figImageNet} shows that subclass specific situations slightly outperform class specific
situations. This is because subdivision into subclasses by clustering yields more specialized object
boundary detectors, which are especially helpful when the object class can occur in different
contexts. Indeed, looking at performance increase of individual classes, the use of subclasses
yields an increase in AP of 0.04, 0.08, and 0.14 for respectively  \emph{killer-whale},
\emph{airship}, and \emph{basketball}. The variety of contexts of the killer-whale can be seen in
Figure~\ref{figSubclassSpecific}, \emph{airships} occur on the ground and against the sky, while
basketball images range from basketball close-ups, to indoor competition (see
Figure~\ref{figExampleImageNet}), to outdoor play.

Note that the monolithic boundary detector is trained exclusively on the objects of interest. Hence
if a local image patch causes a false boundary prediction, it is necessarily similar in appearance
to a local image patch of a true object boundary. Now notice in Figure~\ref{figExampleImageNet} that
monolithic boundary detection fires on many non-object boundary edges: the crowd of the basketball
player, the shade behind the dog, the dog's internal boundaries, and the water of the killer-whale.
Therefore such background edges are necessarily similar in appearance to true object boundaries.
This means a monolithic approach can never work well in all situations.

In contrast, situational object boundary detection performs much better, especially when using
subclass specific situations. On the basketball image, 
our method ignores not only the crowd but also the player, which is good since the player
is not the object of interest. For the dog our method focuses primarily on the dog boundaries
ignoring shadow and its interior boundaries. For the killer-whale spurious edges caused by the water
are ignored.

We conclude that by using object boundary detectors specialized for the identified situation, we
effectively constrain the expected local appearance of object boundaries, which helps resolving
ambiguities.  This yields significant improvements: whereas a monolithic approach results in 0.260
AP, our subclass specific situation yield 0.296 AP, a relative improvement of 14\%.

\mypar{CNNs vs Fisher Vectors.}
Table~\ref{tabImageNet} shows that CNN features work generally better than Fisher vectors for
situational object boundary detection. This confirms other observations on the strength of CNN
features (e.g.~\cite{chatfield14bmvc,donahue13decaf,razavian14cvpr}).
For class-agnostic situations improvements are especially good since it improves both the creation of situations and
the classification. We use CNN features for the remainder of this paper.

\mypar{Using ground-truth image labels.} Table \ref{tabImageNet} includes an experiment where we use
the ground-truth label to determine which class-specific boundary detector should be applied
(\emph{GT - class specific}). This helps assessing the quality of the global image appearance
classifier within our framework.  As the table shows, there is almost no difference between \emph{GT
- class specific} and \emph{CNN - class specific}.  Hence within our framework global image
classification achieves what can be maximally expected from it.

\mypar{CNN features inside the Random Forest.}
Theoretically, the Random Forests can learn from any features of different modalities. So it would
arguably be simpler to directly provide global image features to the Structured Edge
Forests and bypass the intermediate step of classifying images into situations. We
tried this with CNN features, which are stronger and have a lower dimensionality than Fisher
vectors. We name this setting \emph{monolithic - CNN enhanced}.
Table~\ref{tabImageNet} shows that this does not work better than the baseline monolithic detector.

\subsection{Microsoft COCO}

\paragraph{Dataset.}

Microsoft COCO~\cite{lin14eccv} provides accurate segmentations for its 80 object classes such as
\emph{person}, \emph{banana}, \emph{bus}, \emph{cat}, and others.  We use v0.9  consisting of 82,783
training and 40,504 validation images. Images contain on average 7.7 different object classes. Since
evaluation of boundary predictions is relatively slow by necessity~\cite{Martin04}, we limit
evaluation to the first 5,000 images of the validation set (which comes already randomized). 


\mypar{Number of situations.}

For our subclass specific situations, we choose 10 subclasses per class, leading to a total of 800
situations. We also use 800 class agnostic situations.


\mypar{Results.} 

\begin{figure}[t]
        \vspace{-7mm}
    \centering
    \includegraphics[width=0.96\linewidth]{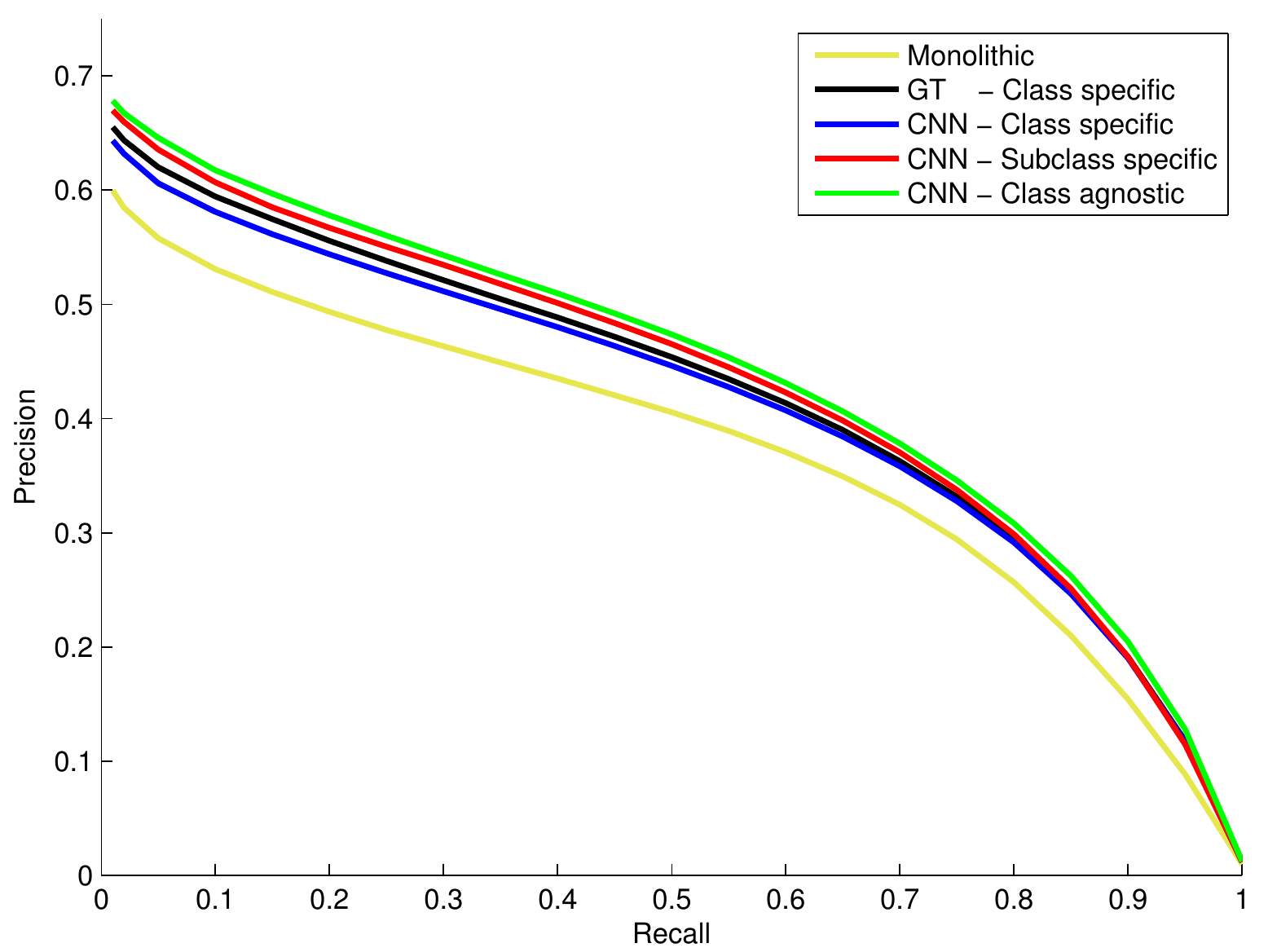}
    \caption{\emph{Performance of object boundary detection on the first 5000 images of the
    COCO validation set.}}
    \label{figCOCOCnn}
        \begin{footnotesize}
    \centering
\begin{tabular}[h]{|l|c|c|c|}
    \hline
    \multicolumn{4}{|l|}{Detectors trained on class object boundaries only}\\
    \hline
    & precision at & precision at & average \\
    & 20\% recall & 50\% recall & precision \\
    \hline
    GT - class specific & 0.566 & 0.460 & 0.422 \\
    CNN   - class specific & 0.543 & 0.443 & 0.407 \\
    CNN   - subclass specific & 0.560 & 0.459 & 0.418 \\
    \hline
    \hline
    \multicolumn{4}{|l|}{Detectors trained on all object boundaries within images}\\
    \hline
    & precision at & precision at & average \\
    & 20\% recall & 50\% recall & precision \\
    \hline
    monolithic & 0.494 & 0.406 & 0.368 \\
    GT - class specific & 0.556 & 0.454 & 0.416 \\
    CNN   - class specific & 0.544 & 0.446 & 0.408 \\
    CNN   - subclass specific & 0.567 & 0.465 & 0.424 \\
    CNN   - class agnostic & 0.578 & 0.474 & 0.434 \\
    \hline
\end{tabular}
\captionof{table}{\emph{Results on Microsoft COCO. Situational object boundary detection significantly
outperforms a monolithic strategy.}}
    \label{tabCOCO}
\end{footnotesize}
\vspace{-4mm}
\end{figure}

In contrast to the previous experiment, here most images contain multiple object classes. Now the
first question is: should we train (sub)class specific object boundary detectors on only the object
boundaries of the target class or on the boundaries of all object classes present in the image?
Results are shown in Table~\ref{tabCOCO}. Interestingly, results are slightly better for true single
class object boundary detectors in the theoretical setting where we use the Ground Truth to
determine the class label (\emph{GT - class specific}). In contrast, when using CNN features results
are slightly better when the detectors are trained on all object boundaries in the images.  This
suggests that mistakes made by object classification can be partially amended by having object
boundary predictors specialized to a certain context rather than to a certain object class.  For the
rest of this paper, we therefore train situational object boundary detectors always on all object
boundaries present in the images of a situation.

Figure~\ref{figCOCOCnn} compares situational object boundary detection with the monolithic baseline.
Whereas a monolithic approach yields an AP of 0.368, our situational approaches yield a substantial
higher AP at 0.408, 0.424, and 0.434 for respectively class specific, subclass specific, and class
agnostic situations.  The best AP improvement is almost 0.07 for class agnostic situations.

As before, subclass specific situations outperform class specific situations. But unlike ImageNet,
on COCO the class agnostic situations slightly outperform the subclass specific. This is likely
because in our ImageNet subset only a single class is annotated, whereas COCO images often
contain multiple classes 
The fact that class agnostic situations are superior suggests that the whole context of the image is
more important for determining which object boundaries to expect than the specific object classes
depicted.

\begin{figure*}[htpb]
    \vspace{-0.7cm}
    \centering
    \includegraphics[width=0.92\linewidth]{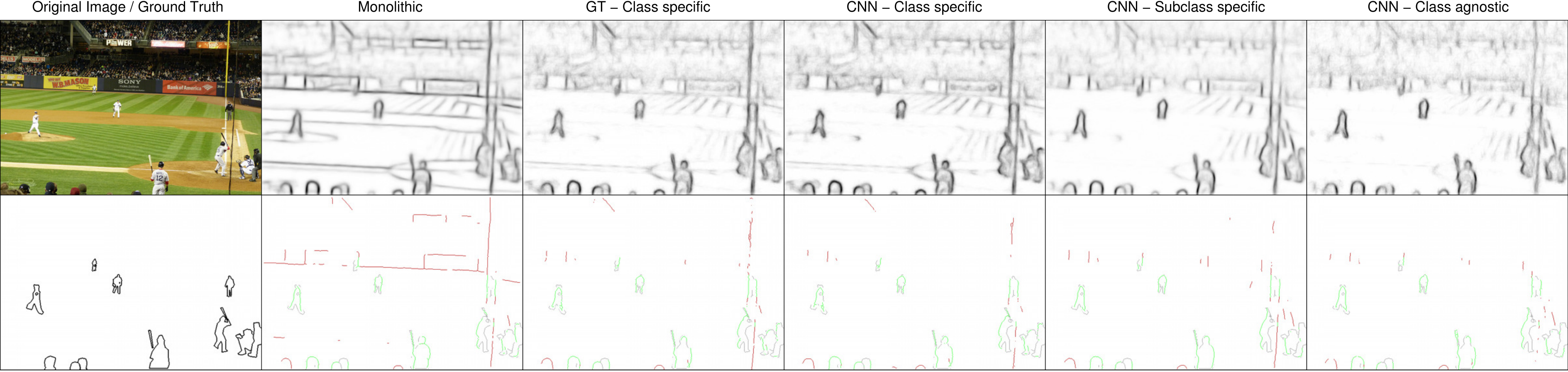}
    \includegraphics[width=0.92\linewidth]{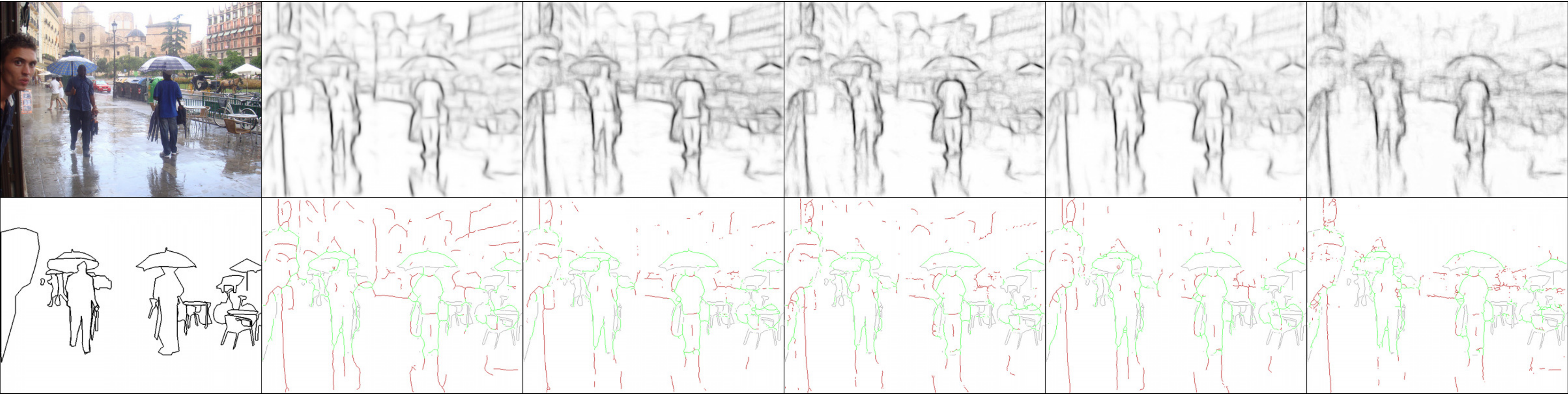}
    \includegraphics[width=0.92\linewidth]{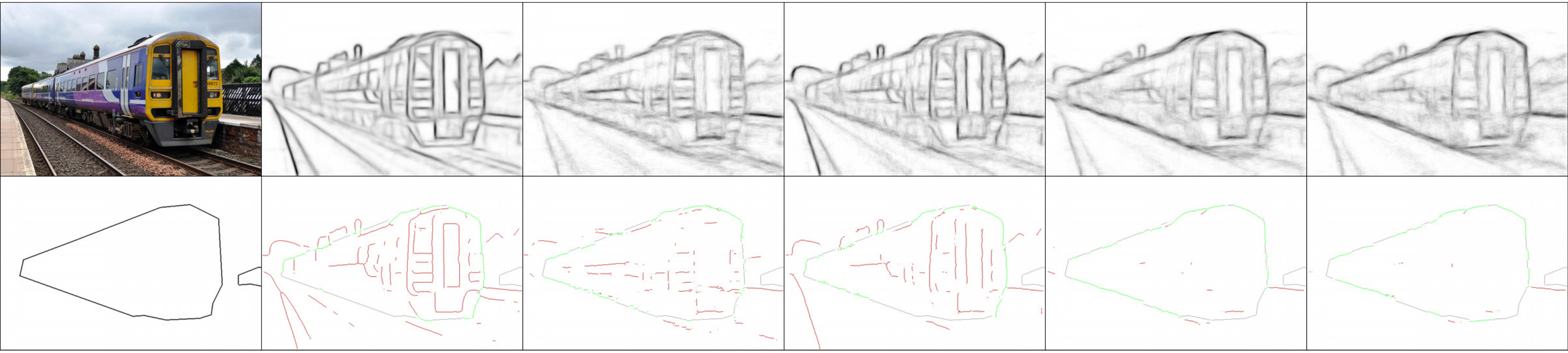}
    \caption{\emph{Examples from COCO. Odd rows: input image and boundary predictions.
        Even rows: ground truth boundaries and precision at a recall of 50\%. True positives are green, false positives red, and
    undetected boundaries grey. While the monolithic 
detector often incorrectly fires on the background and internal boundaries, our situational
object boundary detectors focus better on true object boundaries.}}
    \label{figExampleCOCO}
    \vspace{-5mm}
\end{figure*}

Figure~\ref{figExampleCOCO} shows qualitative results. In contrast to a monolithic approach, our
situational object boundary detector correctly ignores grass/gravel transitions in baseball,
contours of buildings (which are not objects of interest) in streets, and interior boundaries of the
train.

We conclude that by identifying a situation, we can avoid many false positive
object boundary predictions made by a monolithic detector. This leads to significant
improvements: whereas a monolithic approach yields 0.368 AP, class agnostic situations yield 0.434
AP, a relative improvement of 18\%.

\begin{figure}[t]
    \vspace{-5mm}
    \centering
    \includegraphics[width=0.96\linewidth]{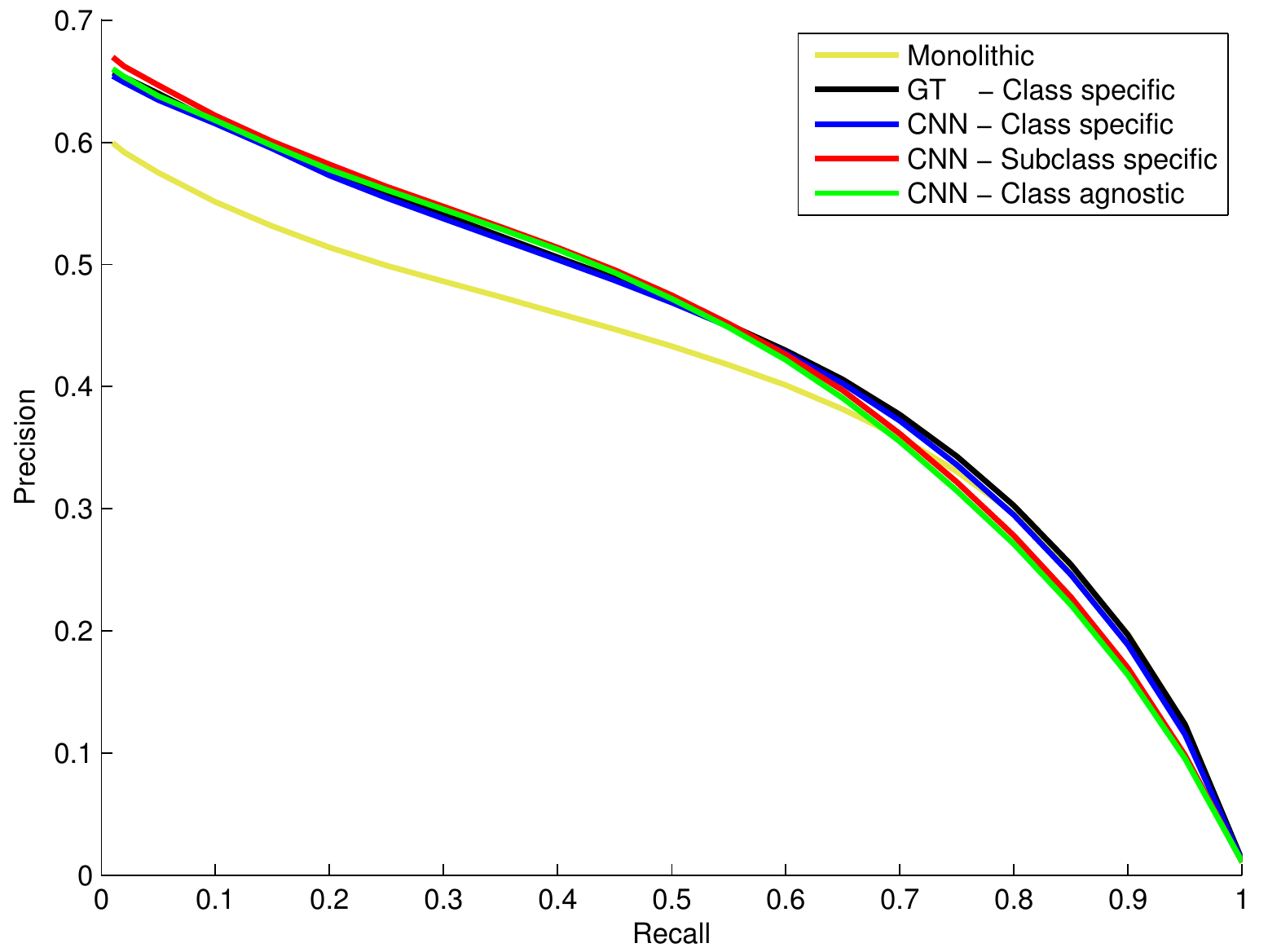}
    \caption{\emph{Performance of object boundary detection on the Pascal VOC 2012
    segmentation database.}}
    \label{figPascalCNN}
    \begin{footnotesize}
\begin{tabular}[hb]{|l|c|c|c|}
    \hline
    & precision at & precision at & average \\
    & 20\% recall & 50\% recall & precision \\
    \hline
    monolithic & 0.514 & 0.433 & 0.396 \\
    GT - class specific & 0.576 & 0.470 & 0.430 \\
    CNN   - class specific & 0.573 & 0.469 & 0.426 \\
    CNN   - subclass specific & 0.582 & 0.475 & 0.426 \\
    CNN   - class agnostic & 0.578 & 0.472 & 0.422 \\
    \hline
\end{tabular}
\captionof{table}{\emph{Results on validation of Pascal VOC 2012 segmentation.}}
    \label{tabPascal}
\end{footnotesize}
\vspace{-4mm}
\end{figure}

\subsection{Pascal VOC 2012 segmentation}\label{cptResultsPascal}

\paragraph{Dataset.} We use the 1,464 training and 1,449 validation images of Pascal VOC 2012
segmentation, annotated with contours for 20 object classes for all instances in all images.

\mypar{Number of situations.} Since the dataset is a lot smaller than Microsoft COCO, we
choose to have 5 subclasses per class to still have sufficient training data per situation, leading
to 100 subclass specific situations. For fair comparison, we also cluster 100 class agnostic
situations. 

\mypar{Results.} Results are presented in Figure~\ref{figPascalCNN} and Table~\ref{tabPascal}.
Again, with 0.426 AP the situational object boundary detection significantly outperforms the
monolithic performance of 0.396 AP. This is a relative 8\% improvement.

On this dataset, class specific situations have about the same performance as subclass specific and
class agnostic. This is different than on ImageNet and COCO, most likely because the training set
is smaller. Hence fine-grained situations yield fewer benefits since both training appearance
based classifiers and training object boundary detectors is more difficult with less data.

\subsection{Semantic Boundaries Dataset (SBD)}\label{cptSemanticContourDetection}

In some applications one may want do `semantic contour detection'~\cite{hariharan11iccv}, i.e. generating class-specific object boundary maps. Our class-specific boundary detectors can produce such maps
$\mathbf{D_c}(I)$, specific to class $c$, using~\eqref{eqMain} but with the summation running only over class $j=c$:
\begin{equation}
    \mathbf{D_c}(I) = P(S_c|I) \cdot D_c(I)
    \label{eqClassSpecific}
\end{equation}
where $P(S_c|I)$ is the probability that class $c$ occurs in image $I$ according to CNN-based
classification. $D_c(I)$ is the output of the class-specific boundary predictor for class $c$.

We use the Semantic Boundaries Dataset of~\cite{hariharan11iccv}, which consists of 11,318 images
from the Pascal VOC 2011 \texttt{trainval} dataset, divided in 8498 training and 2820 test images.
All instances of its 20 object classes were annotated with accurate figure/ground masks by
crowdsourcing. We use the official evaluation software provided by~\cite{hariharan11iccv}.

\begin{figure}[htpb]
    \vspace{-.1cm}
    \centering
    \includegraphics[width=\linewidth]{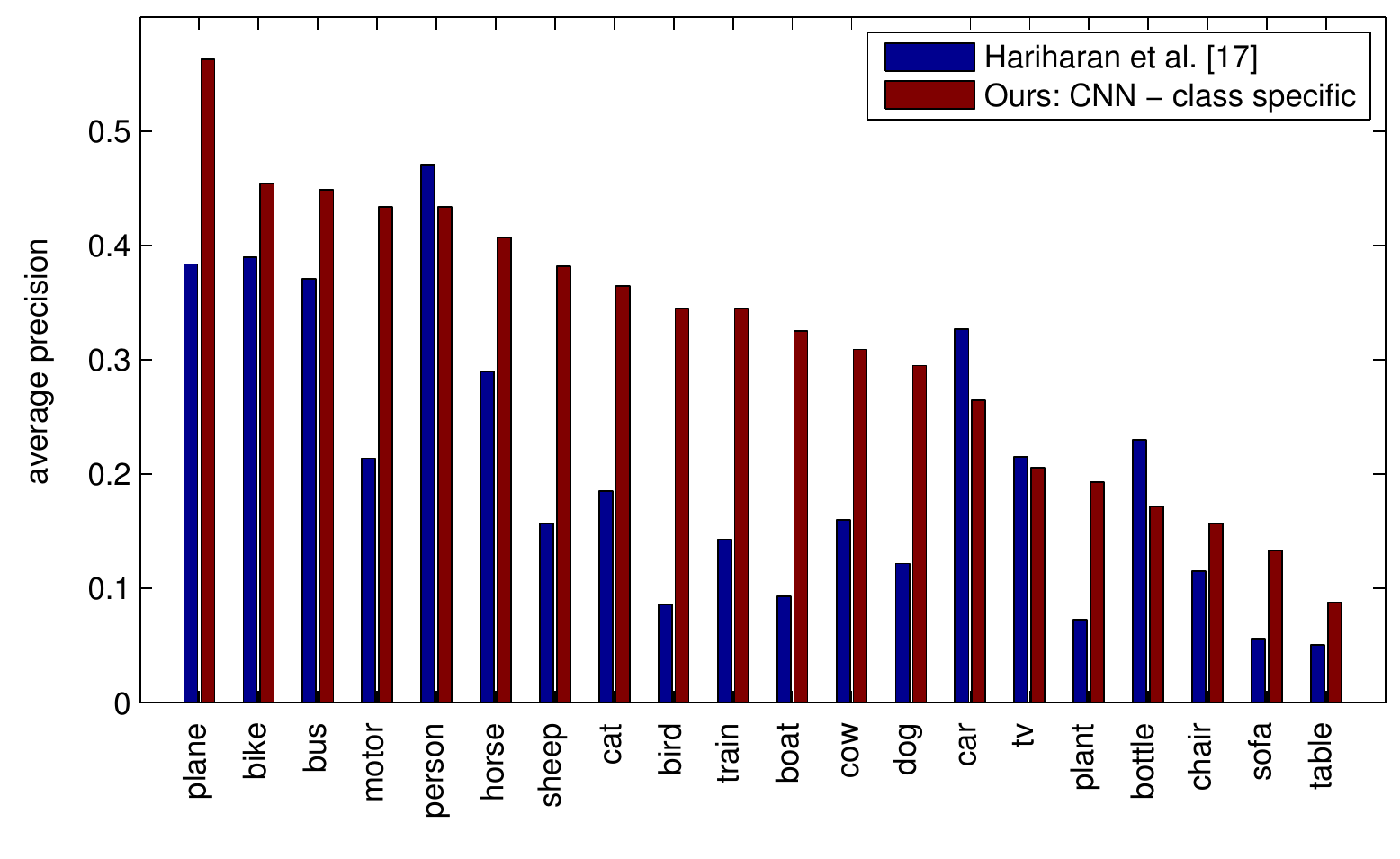}
    \vspace{-6mm}
    \caption{\emph{Semantic Contour Detection on BSD. \cite{hariharan11iccv} versus 
our CNN-based class specific situational object boundary detector.}}
    \label{figHariharan}
    \vspace{-6mm}
\end{figure}

As figure~\ref{figHariharan} shows our method considerably outperforms~\cite{hariharan11iccv} on most classes.
While~\cite{hariharan11iccv} report a mean AP of 0.207, we
obtain 0.316 mAP.

\subsection{Computational Requirements} 

Runtime on a test image is essentially constant for any reasonable number of situations: the most
expensive component is the boundary detector~\cite{Dollar13} which takes 73 ms/image on an
Intel Core i5-3470. At test time we always apply $n=5$ detectors (Equation
\eqref{eqMainCropped}). Extracting CNN features takes about 2 ms/image on a modern GPU. Linear
classification on 4096 dimensions takes less than 2 ms/image for 1000 situations. Hence our
situational object boundary prediction takes around 0.37 s/image, which is still very fast for an object
boundary detector (see e.g.~\cite{Dollar13}).

\section{Conclusion}

The appearance of true object boundaries varies from situation to situation. Hence a monolithic
object boundary prediction approach which predicts object boundaries regardless of the image content
is necessarily suboptimal. Therefore this paper introduces situational object boundary detection.
First the situation is determined based on global image appearance. Afterwards only those boundary
detectors are applied which are specialized for this situation. Since we build on~\cite{Dollar13},
our situational object boundary prediction is fast and takes only 0.37 ms/image.  More importantly,
results on object boundary detection show consistent improvements on three large datasets:
on Pascal VOC 2012 segmentation~\cite{everingham14ijcv}, the automatically segmented
ImageNet~\cite{guillaumin14ijcv,Russakovsky14}, and Microsoft COCO~\cite{lin14eccv}, we obtained
relative improvements of respectively 8\%, 14\% and 18\% AP. Furthermore, on semantic contour
detection our approach substantially outperforms~\cite{hariharan11iccv} on their SBD dataset.

\paragraph{Acknowledgements.}
This work was supported by the ERC Starting Grant VisCul.

{\small
\bibliographystyle{ieee}
\bibliography{jasperRefs,shortstrings,calvin}

\begin{thebibliography}{10}\itemsep=-1pt

\bibitem{Akata14}
Z.~Akata, F.~Perronnin, Z.~Harchaoui, and C.~Schmid.
\newblock Good practice in large-scale learning for image classification.
\newblock {\em TPAMI}, 2014.

\bibitem{Arbelaez11}
P.~Arbel\'{a}ez, M.~Maire, C.~Fowlkes, and J.~Malik.
\newblock {Contour Detection and Hierarchical Image Segmentation}.
\newblock {\em TPAMI}, 2011.

\bibitem{boix12ijcv}
X.~Boix, J.~Gonfaus, J.~van~de Weijer, A.~Bagdanov, J.~Serrat, and J.~Gonzalez.
\newblock Harmony potentials: Fusing global and local scale for semantic image
  segmentation.
\newblock {\em IJCV}, 2012.

\bibitem{bourdev10eccv}
L.~Bourdev, S.~Maji, T.~Brox, and J.~Malik.
\newblock Detecting people using mutually consistent poselet activations.
\newblock In {\em ECCV}, 2010.

\bibitem{Breiman01}
L.~Breiman.
\newblock Random forests.
\newblock {\em Machine learning}, 45(1):5--32, 2001.

\bibitem{Canny86}
J.~Canny.
\newblock A computational approach to edge detection.
\newblock {\em TPAMI}, 1986.

\bibitem{chatfield14bmvc}
K.~Chatfield, K.~Simonyan, A.~Vedaldi, and A.~Zisserman.
\newblock Return of the devil in the details: Delving deep into convolutional
  networks.
\newblock In {\em BMVC}, 2014.

\bibitem{Csurka04}
G.~Csurka, C.~R. Dance, L.~Fan, J.~Willamowski, and C.~Bray.
\newblock {Visual Categorization with Bags of Keypoints}.
\newblock In {\em ECCV International Workshop on Statistical Learning in
  Computer Vision}, Prague, 2004.

\bibitem{Dollar06}
P.~Doll\'ar, Z.~Tu, and S.~Belongie.
\newblock Supervised learning of edges and object boundaries.
\newblock In {\em CVPR}, 2006.

\bibitem{Dollar13}
P.~Doll\'ar and C.~Zitnick.
\newblock Structured forests for fast edge detection.
\newblock In {\em ICCV}, 2013.

\bibitem{donahue13decaf}
J.~Donahue, Y.~Jia, O.~Vinyals, J.~Hoffman, N.~Zhang, E.~Tzeng, and T.~Darrell.
\newblock Decaf: A deep convolutional activation feature for generic visual
  recognition.
\newblock {\em arXiv preprint arXiv:1310.1531}, 2013.

\bibitem{Duda73}
R.~Duda and P.~Hart.
\newblock {\em Pattern Classification and Scene Analysis}.
\newblock 1973.

\bibitem{everingham14ijcv}
M.~Everingham, S.~Eslami, L.~van Gool, C.~Williams, J.~Winn, and A.~Zisserman.
\newblock The pascal visual object classes challenge - a retrospective.
\newblock {\em IJCV}, 2014.

\bibitem{Felzenszwalb10}
P.~F. Felzenszwalb, R.~B. Girshick, D.~McAllester, and D.~Ramanan.
\newblock {Object detection with discriminatively trained part based models}.
\newblock {\em TPAMI}, 2010.

\bibitem{Girshick14}
R.~Girshick, J.~Donahue, T.~Darrell, and J.~Malik.
\newblock Rich feature hierarchies for accurate object detection and semantic
  segmentation.
\newblock In {\em CVPR}, 2014.

\bibitem{guillaumin14ijcv}
M.~Guillaumin, D.~K\"uttel, and V.~Ferrari.
\newblock {ImageNet} auto-annotation with segmentation propagation.
\newblock {\em IJCV}, 2014.

\bibitem{hariharan11iccv}
B.~Hariharan, P.~Arbel\'aez, L.~Bourdev, S.~Maji, and J.~Malik.
\newblock Semantic contours from inverse detectors.
\newblock In {\em ICCV}, 2011.

\bibitem{harzallah09iccv}
H.~Harzallah, F.~Jurie, and C.~Schmid.
\newblock Combining efficient object localization and image classification.
\newblock In {\em ICCV}, 2009.

\bibitem{Jia14}
Y.~Jia, E.~Shelhamer, J.~Donahue, S.~Karayev, J.~Long, R.~Girshick,
  S.~Guadarrama, and T.~Darrell.
\newblock Caffe: Convolutional architecture for fast feature embedding.
\newblock In {\em ACM MM}, 2014.

\bibitem{Jurie05}
F.~Jurie and B.~Triggs.
\newblock {Creating Efficient Codebooks for Visual Recognition}.
\newblock In {\em ICCV}, 2005.

\bibitem{Krizhevsky12}
A.~Krizhevsky, I.~Sutskever, and G.~E. Hinton.
\newblock Imagenet classification with deep convolutional neural networks.
\newblock In {\em NIPS}, 2012.

\bibitem{Lazebnik06}
S.~Lazebnik, C.~Schmid, and J.~Ponce.
\newblock {Beyond Bags of Features: Spatial Pyramid Matching for Recognizing
  Natural Scene Categories}.
\newblock In {\em CVPR}, 2006.

\bibitem{Lim13}
J.~Lim, C.~Zitnick, and P.~Doll\'ar.
\newblock Sketch tokens: A learned mid-level representation for contour and
  object detection.
\newblock In {\em CVPR}, 2013.

\bibitem{lin14eccv}
T.-Y. Lin, M.~Maire, S.~Belongie, J.~Hays, P.~Perona, D.~Ramanan,
  P.~Doll{\'a}r, and C.~Zitnick.
\newblock Microsoft {COCO}: Common objects in context.
\newblock In {\em ECCV}, 2014.

\bibitem{liu09cvpr}
C.~Liu, J.~Yuen, and A.~Torralba.
\newblock {Nonparametric scene parsing: Label transfer via dense scene
  alignment}.
\newblock In {\em CVPR}, 2009.

\bibitem{Lowe04}
D.~G. Lowe.
\newblock {Distinctive Image Features from Scale-Invariant Keypoints}.
\newblock {\em IJCV}, 2004.

\bibitem{Mairal08}
J.~Mairal, M.~Leordeanu, and F.~Bach.
\newblock Discriminative sparse image models for class-specific edge detection
  and image interpretation.
\newblock In {\em ECCV}, 2008.

\bibitem{Marr80}
D.~Marr and E.~Hildreth.
\newblock Theory of edge detection.
\newblock In {\em Proceedings of the Royal Society of London}, 1980.

\bibitem{Martin04}
D.~Martin, C.~Fowlkes, and J.~Malik.
\newblock Learning to detect natural image boundaries using local brightness,
  color, and texture cues.
\newblock {\em TPAMI}, 2004.

\bibitem{Perronnin10}
F.~Perronnin, J.~Sanchez, and T.~Mensink.
\newblock {Improving the Fisher Kernel for Large-Scale Image Classification}.
\newblock In {\em ECCV}, 2010.

\bibitem{Prasad06}
M.~Prasad, A.~Zisserman, and A.~Fitzgibbon.
\newblock Learning class-specific edges for object detection and segmentation.
\newblock {\em Computer Vision, Graphics and Image Processing}, 2006.

\bibitem{Prewitt70}
J.~Prewitt.
\newblock {\em Picture Procesing and Psychopictorics}, chapter Object
  Enhancement and Extraction.
\newblock 1970.

\bibitem{razavian14cvpr}
A.~Razavian, H.~Azizpour, J.~Sullivan, and S.~Carlsson.
\newblock {CNN} features off-the-shelf: An astounding baseline for recognition.
\newblock In {\em DeepVision workshop at CVPR}, 2014.

\bibitem{Roberts65}
L.~Roberts.
\newblock Machine perception of three dimensional solids.
\newblock In {\em Optical and Electro-Optical Information Processing}, 1965.

\bibitem{Russakovsky14}
O.~Russakovsky, J.~Deng, H.~Su, J.~Krause, S.~Satheesh, S.~Ma, Z.~Huang,
  A.~Karpathy, A.~Khosla, M.~Bernstein, A.~Berg, and L.~Fei-Fei.
\newblock Imagenet large scale visual recognition challenge.
\newblock {\em IJCV}, 2015.

\bibitem{Shi00}
J.~Shi and J.~Malik.
\newblock {Normalized Cuts and Image Segmentation}.
\newblock {\em TPAMI}, 2000.

\bibitem{Sivic03}
J.~Sivic and A.~Zisserman.
\newblock {Video Google: A Text Retrieval Approach to Object Matching in
  Videos}.
\newblock In {\em ICCV}, 2003.

\bibitem{torralba03ijcv}
A.~Torralba.
\newblock Contextual priming for object detection.
\newblock {\em IJCV}, 2003.

\bibitem{Uijlings10}
J.~R.~R. Uijlings, A.~W.~M. Smeulders, and R.~J.~H. Scha.
\newblock {Real-time Visual Concept Classification}.
\newblock {\em IEEE Transactions on Multimedia}, 12, 2010.

\bibitem{Vedaldi10VLFeat}
A.~Vedaldi and B.~Fulkerson.
\newblock {VLF}eat - an open and portable library of computer vision
  algorithms.
\newblock In {\em ACM MM}, 2010.

\end{thebibliography}
}

\end{document}